\documentclass[10pt,conference]{IEEEtran}
\usepackage[utf8]{inputenc}
\usepackage{float}
\usepackage[style=ieee,
            sorting=none,
            backend=biber,
            natbib=true,      % allows \citet, \citep as aliases
            giveninits=true,
            maxcitenames=2,   % "Author et al." when >2 authors
            mincitenames=1]{biblatex}
\AtBeginBibliography{%
  \footnotesize            % 8pt font, like IEEEtran BibTeX
  \setlength{\itemsep}{0pt}% no extra space between items
  \setlength{\parskip}{0pt}% no paragraph skip
}
\usepackage{newtxtext,newtxmath} % Times-like text/math
\IEEEoverridecommandlockouts

\usepackage{amsmath,amsfonts}
\usepackage{algorithmic}
\usepackage{graphicx}
\usepackage{textcomp}
\usepackage{xcolor}
\usepackage{listings}
\usepackage{booktabs}
\usepackage{multirow}
\usepackage{balance}
\usepackage{tabularx}
\usepackage{array}
\usepackage{xparse} % for NewDocumentCommand
\usepackage{caption}
\newcolumntype{L}{>{\raggedright\arraybackslash}X}

\usepackage{balance}
\usepackage{listings}
\usepackage{tabularx}
\lstdefinestyle{basiccode}{
    basicstyle=\ttfamily\footnotesize,
    breaklines=true,
    showstringspaces=false,
    frame=single,
    backgroundcolor=\color{gray!10},
    literate={*}{{*}}1,
    xleftmargin=1cm,
    xrightmargin=0pt,
    framexleftmargin=0pt,
    framexrightmargin=0pt,
    framesep=6pt,
    linewidth=0.9\columnwidth   % make box exactly as wide as the column
}
\def\BibTeX{{\rm B\kern-.05em{\sc i\kern-.025em b}\kern-.08em
    T\kern-.1667em\lower.7ex\hbox{E}\kern-.125emX}}

\begin{document}

\title{Evaluating Open-Source Vision-Language Models for
Multimodal Sarcasm Detection\\
}

\makeatletter
\newcommand{\linebreakand}{%
  \end{@IEEEauthorhalign}
  \hfill\mbox{}\par
  \mbox{}\hfill\begin{@IEEEauthorhalign}
}
\makeatother

\author{\IEEEauthorblockN{Saroj Basnet}
\IEEEauthorblockA{\textit{George Mason University} \\
Fairfax, VA, USA \\
%sbasne2@gmu.edu
}
\and 

\IEEEauthorblockN{Shafkat Farabi}
\IEEEauthorblockA{\textit{Virginia Tech} \\
%\textit{Virginia Tech}\\
Alexandria, VA, USA \\
%mfarabi@vt.edu
}
\and
\IEEEauthorblockN{Tharindu Ranasinghe}
\IEEEauthorblockA{\textit{Lancaster University} \\
Lancaster, UK \\
%email address or ORCID
} 

%\and
\linebreakand

\IEEEauthorblockN{Diptesh Kanojia}
\IEEEauthorblockA{\textit{University of Surrey} \\
Guildford, UK\\
%email address or ORCID
}
\and
\IEEEauthorblockN{Marcos Zampieri}
\IEEEauthorblockA{\textit{George Mason University} \\
Fairfax, VA, USA \\
%email address or ORCID
}
% \and
}

\maketitle

\begin{abstract}
Recent advances in open‑source vision‑language models (VLMs) offer new opportunities for understanding complex and subjective multimodal phenomena such as sarcasm. In this work, we evaluate seven state‑of‑the‑art VLMs - BLIP2, InstructBLIP, OpenFlamingo, LLaVA, PaliGemma, Gemma3, and Qwen‑VL - on their ability to detect multimodal sarcasm using zero‑, one‑, and few‑shot prompting. Furthermore, we evaluate the models' capabilities in generating explanations to sarcastic instances. We evaluate the capabilities of VLMs on three benchmark sarcasm datasets (Muse, MMSD2.0, and SarcNet). Our primary objectives are twofold: (1) to quantify each model’s performance in detecting sarcastic image–caption pairs, and (2) to assess their ability to generate human‑quality explanations that highlight the visual–textual incongruities driving sarcasm. Our results indicate that, while current models achieve moderate success in binary sarcasm detection, they are still not able to generate high-quality explanations without task-specific fine-tuning.
\end{abstract}


\section{Introduction}

Sarcasm is a nuanced form of communication where the intended meaning diverges from the literal expression conveying irony or mockery (see Figure \ref{fig:example_vis}). Automatic sarcasm detection has been widely studied in texts \cite{info13080399,helal2024contextual,yaghoobian2021sarcasmdetectioncomparativestudy} and it is an important part of subjective language understanding \cite{song2025large} along with tasks such as sentiment analysis \cite{dmonte2024evaluation} and offensive language identification \cite{ranasinghe2020multilingual,ranasinghe2021mudes}. Yet, on social media, sarcasm is often achieved by pairing text with images, and the ironic effect arises from the mismatch between the visual and textual objects. Multimodal sarcasm detection (MSD) has attracted increasing attention due to the proliferation of multimedia content on social platforms. A recent comprehensive survey on the topic \cite{mfarabi} highlights the complexity of MSD and the necessity for models capable of interpreting cross-modal incongruities.

Vision–language models (VLMs) pre-trained on large image–text corpora exhibit powerful zero‑ and few‑shot abilities across tasks \cite{yang2023revilmretrievalaugmentedvisuallanguage}. 
%and in‑context learning lets them adapt to new tasks via prompting alone \cite{yang2023revilmretrievalaugmentedvisuallanguage}. 
The state-of-the-art vision-language models (VLMs) have shown promise in handling multimodal data. Models like Flamingo \cite{NEURIPS2022_960a172b} and VILA \cite{lin2024vilapretrainingvisuallanguage} have demonstrated capabilities in zero-shot and few-shot learning, adapting to new tasks with minimal examples through in-context learning. These models process interleaved sequences of visual and textual data, making them suitable candidates MSD. However, their performance on this task remains largely unexplored. Notably, studies like  \citet{lin2023goatbench} and \citet{yang2023mm} - include MSD within 14 broader multimodal benchmark datasets used to evaluate VLMs. While these studies benchmark models like GPT-4, LLaMA, OpenFlamingo, BLIP2, and InstructBLIP, only minimal sarcasm data is present. 

\begin{figure}[!ht]
    \centering
    \includegraphics[width=0.75\columnwidth]{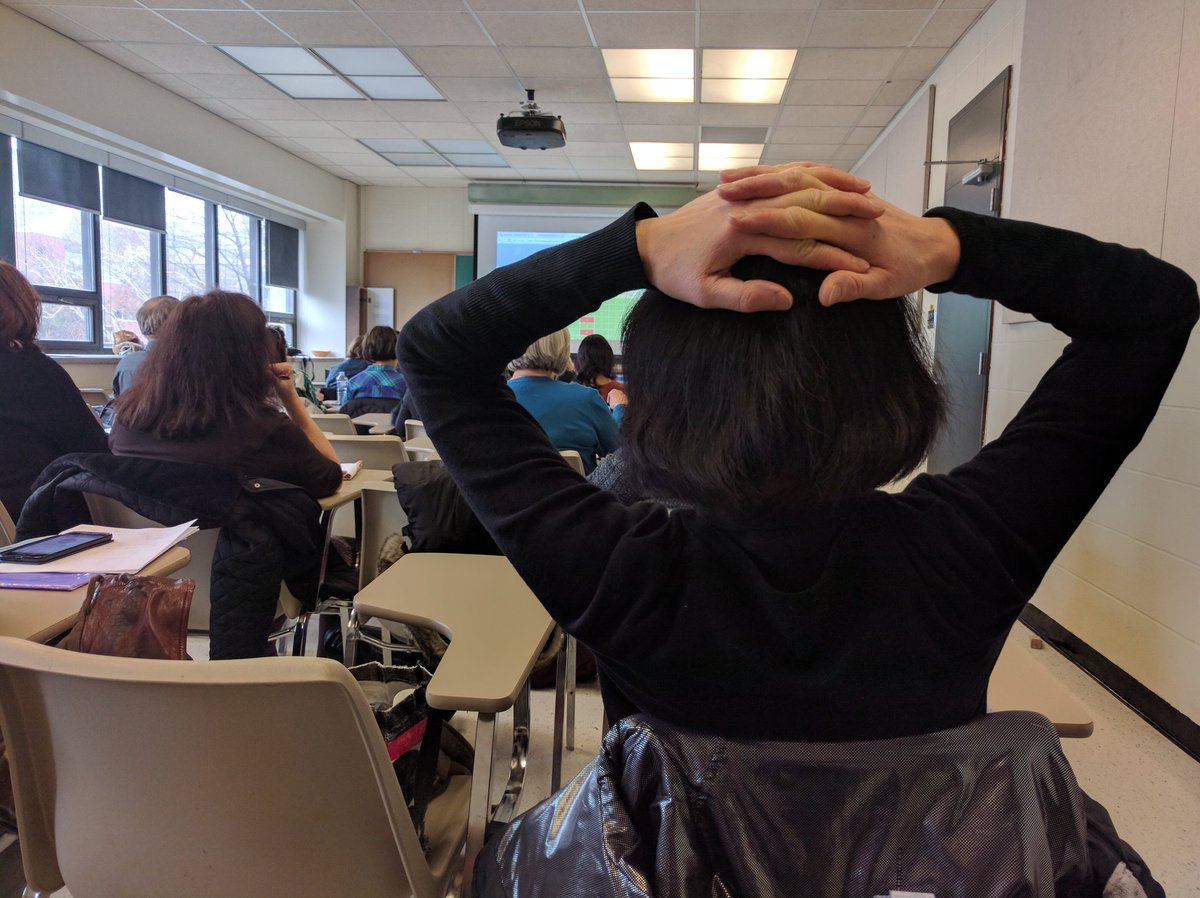} \\
   {\bf wow. what a great view!}
    \caption{Example of a sarcastic instance (image and text pair) from the dataset.}
    \label{fig:example_vis}
\end{figure}

\noindent In contrast, \cite{tang-etal-2024-leveraging} have incorporated multimodal large language models (MLLMs) in their pipeline, utilizing CLIP and other visual encoders with training to achieve state-of-the-art performance on MSD. Given the nuanced nature of sarcasm and its reliance on context across modalities, it is imperative to assess whether these models can detect and explain sarcasm without extensive fine-tuning \cite{mfarabi,ijcai2024p252}. Recent work on Multimodal Sarcasm Explanation (MuSE) \cite{desai2021nice} requires models to generate human‑style explanations for sarcastic image–caption pairs, but relies on supervised fine‑tuning and specialized architectures. We instead ask: \emph{Can off‑the‑shelf open‑source VLMs detect and explain multimodal sarcasm purely through in‑context prompts?}

Building on this foundation, our study focuses on benchmarking widely used open-source VLMs for their ability to identify sarcasm without fine-tuning, utilizing several datasets curated specifically for MSD. Beyond classification, we also evaluate their capability of generating coherent textual explanations reasoning solely through prompting. 

The paper addresses the following research questions (RQs): \\

\begin{itemize}
\item {\bf RQ1:} How effectively do state-of-the-art open-source VLMs perform on MSD benchmarks using in-context learning?
\item  {\bf RQ2:} Can state-of-the-art open-source VLMs generate coherent and task-relevant textual explanations for multimodal sarcasm reasoning solely through prompting?
\end{itemize}

\vspace{2mm}

\noindent The main contributions of this paper are as follows: \\

    \begin{enumerate}
        \item We provide a unified in‑context framework with prompt templates that integrate images, few‑shot examples, and explanation seeds across seven VLMs.
        \item We carry out a classification benchmark with systematic zero‑, one‑, few‑shot evaluation on three MSD benchmark datasets (MMSD, MMSD2.0, SarcNet).
        \item We evaluate the explanation generation capabilities each model’s free‑form irony explanation generation.
        % \item Practical guidelines - Insights for the limits of in‑context multimodal reasoning, and paths for fine‑tuning and improved evaluation.
    \end{enumerate}

%\vspace{-2mm}

\section{Related Work}

%\vspace{-2mm}

Automatic sarcasm detection has been extensively studied due to its challenging nature \cite{chaudhari2017literature,srirag2024besstie}. Furthermore, sarcasm has been studied as a linguistic phenomenon part of various related NLP tasks such as sentiment analysis \cite{joshi2017automatic,badlani2019ensemble,maynard2014cares}, offensive speech identification \cite{frenda2018role}, and humor identification \cite{rothermich2021change}. Most research on this task addresses the identification of sarcasm in texts \cite{chaudhari2017literature,verma2021techniques,salini2023sarcasm}. 

\citet{schifanella2016detecting} demonstrate that visual modalities often include cues that help models identifying sarcastic intent and since then multiple studies have addressed MSD. Multiple MSD datasets have been crated from social media data \cite{schifanella2016detecting,das2018sarcasmB,cai2019multi}. Among these, MMSD \cite{cai2019multi} and its improved version MMSD2.0 \cite{qin-etal-2023-mmsd2} have become de-facto benchmarks for MSD. \citet{yue-etal-2024-sarcnet} recently propose a multilingual multimodal dataset for MSD named SarcNet. 

Earlier work on MSD focused on extracting visual and textual features with separate encoders (BERT, ResNet, VIT etc.) and designing techniques to aggregate these features. 
%to train a sarcasm classifier on. 
Most feature aggregation techniques rely on the fact that sarcasm is marked by a characteristic semantic incongruity between the text and image. Various techniques such as inter-modal attention \cite{pan2020modeling}, graph neural nets \cite{liang2022multib,liang2021multia}, and D\&R network \cite{xu2020reasoning} have been employed to model this relationship at the feature aggregation stage. More recent works have shifted their focus to model the interaction between the modalities early on in the processing pipeline. For example, \citet{wang-etal-2020-building} benchmark multimodal models like VisualBERT, LXMERT, ViLBERT etc. that jointly extract features from images and texts to identify sarcasm. \citet{qin-etal-2023-mmsd2} also take this route by using OpenAI's CLIP architecture.  

In recent years, multimodal LLMs such as GPT-4, LLaVA, and others have demonstrated remarkable capabilities in extracting information and analyzing sentiment from multimodal data. These models have achieved state-of-the-art results across a variety of multimodal tasks. However, as noted by \citet{mfarabi} in their survey of the field, there remains a significant gap in the literature: LLMs are still underexplored for MSD. This work helps filling this gap presenting a systematic evaluation of open-source LLMs on MSD using in-context learning. 

% Only a few studies—such as \citet{lin2023goatbench} and \citet{yang2023mm}—include MSD within broader multimodal benchmark datasets used to evaluate visual-language models (VLLMs). While these works benchmark models like GPT-4, LLaMA, OpenFlamingo, BLIP, and InstructBLIP, sarcasm detection is treated as a peripheral task and minimal data pertaining to MSD is present. \cite{tang-etal-2024-leveraging} incorporate multimodal LLMs in their pipeline together with CLIP and other visual encoders with training to achieve state-of-the-art performance on MSD. 

% In contrast to these works, we focus on benchmarking widely used and general purpose multimodal LLMs ability to identify sarcasm with no fine tuning on several large-scale datasets curated solely for MSD. Beyond classification, we also evaluate the ability of VLLMs to generate sarcastic content in multimodal contexts. Although \citet{zhao-etal-2023-multi} also explore sarcasm generation, their approach relies on training multimodal transformers with supervised learning. In our study, we assess the zero-shot and few-shot capabilities of general-purpose VLLMs for this task.

%TODO: Half a page RW building on our survey. 
\section{Methods}

\paragraph{Datasets}

We utilize three publicly available datasets presented in Table \ref{tab:dataset_comparison} for multimodal sarcasm detection. For evaluation, we randomly sample 3,000 image-caption pairs from both MMSD2.0 and SarcNet, multilingual - English and Chinese; datasets for binary classification. In parallel, we draw explanations from the Muse dataset to assess the models' ability not only to label but to qualitatively justify their predictions. 

\begin{itemize}
    \item \textbf{MuSE} \cite{desai2021nice}: Contains image-text pairs labeled for sarcasm, notably including explanations for sarcastic instances. Primarily used for positive (sarcastic) examples in our few-shot prompts. 
    \item \textbf{MMSD2.0} \cite{qin-etal-2023-mmsd2}: A comparatively larger dataset for multimodal sarcasm detection featuring both sarcastic and non-sarcastic examples. Used for sourcing negative (non-sarcastic) examples for few-shot prompts.
    \item \textbf{SarcNet} \cite{yue-etal-2024-sarcnet}: SarcNet is a multilingual (English and Chinese) dataset for multimodal sarcasm detection, comprising 3,335 image-text pair samples. 
  \end{itemize}

\begin{table}[!ht]
  \centering
  \caption{Overview of key characteristics of datasets used in multimodal sarcasm detection, highlighting the distribution of positive (+) and negative (-) examples, availability of explanations, and support for multilingual data.}
  \label{tab:dataset_comparison}
  \begin{tabularx}{\columnwidth}{lXXXX}
    \toprule
    \textbf{Dataset} & \textbf{+ examples} & \textbf{- examples} & \textbf{Explanation} & \textbf{Multilingual} \\
    \midrule
    Muse      & 3,510 & 0 & $\checkmark$ & $\times$ \\
    MMSD2.0   & 11,651 & 12,980 & $\times$ & $\times$ \\
    SarcNet   & 1,875 & 1,460 & $\times$ & $\checkmark$ \\
    \bottomrule
  \end{tabularx}
\end{table}

\paragraph{Models} We benchmark seven open‑source VLMs spanning diverse architectures and pretraining regimes:

\begin{itemize}
    \item \textbf{InstructBlip} \cite{dai2023instructblip} - FlanT5 large model further instruction‑tuned on multi‑task vision‑language data.
    \item \textbf{Blip2} \texttt{2.7B} \cite{li2023blip2} - frozen image encoder with Q‑former and a large language model backbone.
    \item \textbf{OpenFlamingo} \texttt{3B} \cite{awadalla2023openflamingoopensourceframeworktraining} - a lightweight adaptation of Flamingo for open‑source environments.
    \item \textbf{LLaVA} \texttt{7B} \cite{liu2023llava} - vision‑language alignment via adversarial fine‑tuning of LLaMA.
    \item \textbf{PaliGemma} \texttt{3B} \cite{beyer2024paligemma} - multimodal mixture‑of‑experts model combining visual and textual experts.
    \item \textbf{Qwen‑VL} \texttt{7B} \cite{Qwen-VL} – Q‑aware encoder‑decoder with unified multimodal attention.
    \item \textbf{Gemma3} \texttt{27B} \cite{gemma_2025}–  instruction-tuned multimodal model that processes both images and text, and broad multilingual support. 
\end{itemize}

\paragraph{In‑Context Prompting}

We apply a unified \texttt{global} prompt instruction template across all-shots settings to ensure the consistency in how each model is queried: %The unified prompt template elicit both classification and explanation from each VLM

\begin{lstlisting}[style=basiccode, caption={Sarcasm Detection Prompt Structure}, label={lst:prompt_md_single}]
*<global_instruction>*
Example: (zero-, one-, few-shots) 
*<image>*
*Caption:<caption> Answer: Yes/No*
*<image>*
**Context:**: {caption}"
Is this sarcastic?
\end{lstlisting}

\begin{itemize}
    \item \texttt{Zero-shot}: no examples or reference explanations included.
    \item \texttt{One‑shot / Few‑shot}: we prepend 1 $-$ 3 exemplar triplets (caption, reference explanation, \texttt{“Yes.”} or  \texttt{“No.”}) drawn from Muse (positive) and MMSD2.0 (negative).
\end{itemize}

\noindent Prompts are truncated/padded to each model’s maximum token length; image inputs use each model’s native preprocessing pipeline (PIL Image pixel‑values or embedded file‑paths). However, due to Gemma3’s maximum context window, including more than one example in a few-shot prompt would exceed its capacity, so we omit few-shot evaluations for this model.

\begin{lstlisting}[style=basiccode, caption={Explanation Generation Prompt Structure}, label={lst:expl_lst}]
*<Context>:*
*<image>*
**Original Caption**: {caption}"
**Provided Explanation**: {explnation}
**Task Instruction**
\end{lstlisting}

\noindent This study utilized zero-, one-, and few-shot prompting strategies to evaluate Multimodal sarcasm detection. Zero-shot prompting provided only task instructions and the target query, relying on the model's pre-trained or tuned knowledge. One-shot and few-shot approaches provided one or multiple (N= 1, 3 pairs used for asset building) complete examples, respectively, within the prompt to enable in-context learning. All strategies aimed for a unified output (\texttt{Yes/No} classification). Furthermore, we prompted each model to generate a sarcasm explanation using the template in Listing \ref{lst:expl_lst}. All explanation generation was performed in a strictly zero-shot setting.

\paragraph{Reproducibility} All prompt-templates, random-seeds, and data splits are released publicly on Github.\footnote{\url{https://github.com/pvsnp9/sarcasm_bench}}
Model checkpoints and hyper-parameters are also logged for ease of reproducibility. For further details, see Appendix \ref{app:exp}

%We release all prompt templates, random seeds, and data splits on GitHub \cite{pvsnp9_profile_repo_2025}. Model checkpoints and hyper-parameters (max lengths, beam sizes) are logged alongside results. This ensures that both classification and explanation experiments can be exactly reproduced by other researchers. For further details on the experiment, see Appendix \ref{app:exp}.

\section{Results}

All models were evaluated \texttt{as is} with no finetuning or gradient updates. This ensures a fair, out-of-box comparison under strict inference settings. We employed the prompt strategy presented in the previous section. Any generation that failed to adhere precisely to the prescribed \texttt{Yes} / \texttt{No} format. 

%One-shot and few-shot prompting yielded minimal or negative gains because sarcasm detection requires more than pattern matching from a small number of demonstrations. Even well‐chosen examples cannot fully convey the deep world knowledge and fine‐grained image–text relationships that underlie ironic intent.  Sometimes, adding a single or few demonstrations often introduces noise rather than informative context. The models struggle to generalize the nuanced contradiction from just one example, and strict context‐window limits force truncation of prompt content. Consequently, simply appending additional shots fails to provide sufficient signal to improve performance on this inherently challenging task.

% \begin{figure*}[h!]
%     \centering
%     \caption{Explanation scores for VLM} 
%     \includegraphics[width=1.0\textwidth]{imgs/exp.png}
    
%     \label{fig:exp_score} 
% \end{figure*}

\begin{table}[htb]
  \centering
  \caption{Classification accuracy results of models on the two datasets in one-shot, zero-shot, and few-shot settings. Model and prompting combination sorted by performance.}
  \label{tab:results_generated}
  \scalebox{.9}{ 
  \begin{tabular}{l l l c}
    \toprule
    \textbf{Dataset} & \textbf{Model} & \textbf{Inference Type} & \textbf{Accuracy} \\
    \midrule

    \multirow{19}{*}{SarcNet}
        & Gemma3       & One-shot  & 0.67 \\
        & InstructBlip & Zero-shot & 0.67 \\
        & InstructBlip & Few-shot  & 0.66 \\
        & Gemma3       & Zero-shot & 0.66 \\
        & Qwen         & Few-shot  & 0.59 \\
        & Qwen         & One-shot  & 0.58 \\
        & Paligemma    & Zero-shot & 0.57 \\
        & Paligemma    & One-shot  & 0.57 \\
        & LlaVA        & Zero-shot & 0.57 \\
        & LlaVA        & One-shot  & 0.56 \\
        & Paligemma    & Few-shot  & 0.55 \\
        & Openflamingo & One-shot  & 0.64 \\
        & Openflamingo & Few-shot  & 0.56 \\
        & Openflamingo & Zero-shot & 0.43 \\
        & Blip2        & Few-shot  & 0.44 \\
        & Blip2        & Zero-shot & 0.43 \\
        & Blip2        & One-shot  & 0.43 \\
        & Qwen         & Zero-shot & 0.43 \\
        & LlaVA        & Few-shot  & 0.43 \\

    \midrule

    \multirow{20}{*}{MMSD2}
        & Gemma3       & One-shot  & 0.73 \\
        & InstructBlip & Zero-shot & 0.64 \\
        & InstructBlip & Few-shot  & 0.63 \\
        & InstructBlip & One-shot  & 0.63 \\
        & Paligemma    & One-shot  & 0.57 \\
        & Paligemma    & Zero-shot & 0.56 \\
        & LlaVA        & Few-shot  & 0.51 \\
        & Qwen         & One-shot  & 0.51 \\
        & Qwen         & Zero-shot & 0.51 \\
        & Qwen         & Few-shot  & 0.50 \\
        & Openflamingo & Few-shot  & 0.49 \\
        & LlaVA        & One-shot  & 0.48 \\
        & Paligemma    & Few-shot  & 0.48 \\
        & Openflamingo & One-shot  & 0.47 \\
        & LlaVA        & Zero-shot & 0.46 \\
        & Blip2        & Few-shot  & 0.53 \\
        & Blip2        & Zero-shot & 0.53 \\
        & Blip2        & One-shot  & 0.52 \\
        & Openflamingo & Zero-shot & 0.52 \\
        & Gemma3       & Zero-shot & 0.60 \\

    \bottomrule
  \end{tabular}
  } % end scalebox
\end{table}

\subsection{Classification Results}

Table \ref{tab:results_generated} presents the results of all models and prompting strategies on the two datasets (SarcNet and MMSD2.0) in terms of accuracy. The table reveals a clear performance spectrum across the seven multimodal models we evaluated on the two datasets under zero, one, and few-shot prompting. Overall, Gemma3 consistently achieved the strongest result with 0.67 accuracy with one-shot prompting on SarcNet and maintained robust performance across MMSD2.0 with 0.73. Instruction tuned models such as InstuctionBLIP also performed competitively, attaining 0.67, on SarcNet dataset and 0.64 on MMSD2.0 dataset. 

%We observe that one-shot and few-shot prompting yielded minimal or negative gains because of the nature of the phenomenon. 

We observe that few-shot prompting does not bring performance improvements over one-shot and few-shot prompting strategies. As discussed a recent survey \cite{mfarabi}, sarcasm detection requires more than pattern matching from a small number of demonstrations. Even well‐chosen, carefully-curated examples cannot fully convey the deep world knowledge and fine‐grained image–text relationships that underlie ironic intent. Sometimes, adding a single or few demonstrations often introduces noise rather than informative context in the prompts. The models struggle to generalize the nuanced contradiction from just one or a few examples example, and strict context‐window limits force truncation of prompt content. Consequently, simply appending additional shots fails to provide sufficient signal to improve performance on this inherently challenging task.

%Qwen demonstrated solid one-shot gains, achieving 0.59 accuracy on SracNet and 0.51 on MMSD2.0, highlighting its ability to benefit from minimal demonstration. In contrast, Blip2 consistently underperformed with accuracies between 0.43, and 0.53 across both datasets, showing minimal benefit from in-context prompting. Collectively, these results underscore that while Gemma3 leads in reliability and overall accuracy, instruction-tuned models deliver meaningful improvements, whereas generic pretrained models remain less effective for multimodal sarcasm detection.

%Table \ref{tab:results_generated} reveals a clear performance spectrum between our evaluated models. Among these models, Gemma3 achieves the highst accuracy more consistantly than other models (SarcNet one-shot prompting and MMSD2.0 zero, one-shot prompting). Instruction tuned models clearly had an advantage over other models, achieving higher accuracies. Qwen demonstrated solid one-shot gains,  highlighting its ability to benefit from minimal demonstration. Blip2 consistently underperformed showing minimal benefit from in-context prompting. Collectively, these results underscore that while Gemma3 leads in reliability and overall accuracy, instruction-tuned models deliver meaningful improvements, whereas generic pretrained models remain less effective for multimodal sarcasm detection.

Overall, while generic vision-language models struggle with the subtlety of sarcasm, instruction tuned architectures like Gemma3 and InstructionBLIP stand out for their robust zero- and one-shot performance. These results highlight the importance of model pretraining, instruction-tuning, and prompt design when tackling high-level reasoning tasks such as multimodal sarcasm detection.

\subsection{Explainability: Generating Explanations}

%To evaluate explanation quality, we employed four complementary metrics spanning different aspects of alignment. Surface-level overlap was assessed using BLEU-4, ROUGE-L, and METEOR, which measure how closely the generated wording matches the human-written explanation. Semantic adequacy was captured via BERTScore-F1, which uses contextual embeddings to assess paraphrase and nuance. For cross-modal faithfulness, we used $\Delta$-CLIPScore to quantify the improvement in CLIP alignment from caption to explanation, indicating how well the explanation bridges the visual–textual irony gap. Finally, VQAScore was used to assess task-driven consistency by verifying whether a VQA model’s answers to questions derived from the explanation matched the image content.
We compute CLIPScore \cite{radford2021clip,hessel2021clipscore} to assess image-text alignment. We evaluate $\Delta$-CLIPScore to quantify the improvement in CLIP alignment from caption to explanation. We introduce the delta CLIP score ($\Delta$CLIP) defined as:
$$
\Delta \text{CLIP} = \text{CLIP}(IMG, G_{\text{exp}}) - \text{CLIP}(IMG, B_{\text{exp}})
$$
where $G_{exp}$ models explanation, and $B_{exp}$ dataset explanation. 
Higher $\Delta$ CLIP score indicates stronger improvements in grounding. This follows the general idea of "delta" comparative metrics used in other multimodal contexts \cite{hertz2023delta} .

We computed the score over 3,000 samples and present the results in Table \ref{tab:clipscore}. Both mean and variance are reported to reflect central tendency and stability across the evaluation set. 

When prompted to articulate the visual–textual incongruity on the Muse dataset, the seven VLMs exhibit widely varying explanation quality. %Figure \ref{fig:exp_score} 
%Table \ref{tab:clipscore} presents that, 
Even on a dataset of exclusively sarcastic examples with explanations, most vision–language models struggle to reproduce those rationales. Among the seven models we evaluated, Salesforce’s BLIP2 achieves the second best overlap with the reference text and a positive embedding-level similarity to the image, making it the clear leader. Instruction-tuned variants like InstructBLIP and OpenFlamingo occupy a middle tier: they generate plausible, partially overlapping explanations but display mixed performance in grounding those explanations back to the image.

\begin{table}[htbp]
\centering
\caption{Mean and Variance of $\Delta$-CLIPScore Across Models for Explanation Generation}
\begin{tabular}{lrr}
\hline
\textbf{Model} & \textbf{Mean} & \textbf{Variance} \\
\hline
    Llava & 1.966 & 27.315 \\     
    Blip2 & 0.831 & 25.532 \\
    Paligemma & 0.757 & 16.234 \\
    Instructblip & 0.583 & 27.749 \\
    Gemma3 & -2.063 & 46.481 \\
    OpenFlamingo & -1.750 & 11.526 \\
    Qwen & -7.143 & 25.515 \\
\hline
\end{tabular}
\label{tab:clipscore}
\end{table}

\noindent LLaVA stands out for producing the most visually grounded explanations, as measured by CLIP alignment, even though its raw text-overlap scores remain modest. At the other extreme, both Qwen-VL and Paligemma barely exceed a random baseline in any metric, and Google’s Gemma3 fails to generate any measurable overlap or alignment with the reference explanations. Finally, every model scores zero on our VQA-style metric, underscoring that none can consistently answer a direct “why” question about the image–caption contrast. These results highlight that only specialized architectures like BLIP2 and LLaVA can approximate human-style sarcasm explanations without further fine-tuning.

While Gemma3 achieves the strongest zero- and one-shot classification accuracy, yet it fails entirely on our text-overlap and embedding-alignment metrics for explanations. Conversely, Llava, one of the weakest classifiers, failing to detect sarcasm, nevertheless generates the most faithful, human-like explanations. Qwen-VL, despite its strong in-context classification performance, produces essentially no overlap with reference explanations. 

Finally, we note that the models that excel at binary sarcasm classification are not the same ones that produce high‐quality explanations. Models like Gemma3 and Qwen-VL are instructed and tuned primarily for discriminative ``Yes/No" decisions, and chat-style generation, so they excel at binary classification. In contrast, high-quality explanations require a generative, VQA-style architecture with deep image–text grounding. Thus, it is the training objective and model design, not only size, that drive this divergence: classification-focused models label effectively, whereas VQA-tuned models produce more faithful, image-grounded rationales.  This divergence underscores that task-specific fine-tuning or architectural adaptations may be needed to bridge classification accuracy and explanatory quality in multimodal sarcasm.

\section{Conclusion and Future Work}

This study provides an evaluation of seven open-source VLMs for MSD and multimodal sarcasm explanation generation using in-context learning. Our experiments addressed two core research questions: {\bf RQ1:} How effectively do state-of-the-art open-source VLMs perform on MSD benchmarks using in-context learning? and {\bf RQ2:} Can state-of-the-art open-source VLMs generate coherent and task-relevant textual explanations for multimodal sarcasm reasoning solely through prompting?

Our findings reveal that, instruction-tuned models such as Gemma3 demonstrate strong performance in binary sarcasm classification; particularly in zero- and one-shot scenarios - answering {\bf (RQ1)}. Our findings also indicate that none of the evaluated models achieves both high classification accuracy and high-quality explanation generation within the same architecture. Models like BLIP2 and LLaVA, despite their lower classification scores, consistently generate explanations that more closely align with human-written rationales. This divergence underscores a critical challenge in multimodal sarcasm: the ability to discriminate sarcasm does not necessarily translate to the capacity for nuanced, interpretable explanation - answering {\bf (RQ2)}.

We observe that increasing the number of in-context examples (few-shot) does not yield gains in sarcasm detection performance. This suggests that current VLMs may require more than simple demonstration-based prompting to generalize the intricate patterns of multimodal irony. In explanation generation, models frequently struggle with both the surface form and semantic adequacy, highlighting the gap between discriminative and generative capabilities in existing architectures.

In future work, we will focus on bridging the gap between classification and explanation. This includes developing hybrid training objectives that jointly optimize for both tasks (e.g., multi-task learning \cite{zampieri2023offensive}). We are experimenting with chain-of-thought (CoT) and multi-stage prompting to elicit richer model reasoning, and fine-tuning on datasets that combine sarcasm detection with human-written explanations. Extending model context windows and integrating RAG techniques or external knowledge could further enhance the models’ ability to capture the subtle contextual cues inherent in sarcasm. Finally, we would also like to expand benchmarks to include adversarial and multilingual examples, as well as incorporating human-in-the-loop evaluations.

% \section*{Ethics Statement}
% %I am not sure if we need to keep it or !!

% In this paper we used publicly-available benchmark datasets. No new data collection has been carried out as part of this work. We did not collect or process writers’/users’ information. We also have not processed users' timelines or carried out any form of user profiling protecting users' privacy and anonymity. All classification and generation experiments have been performed on the basis of individual posts. 

%\balance

\printbibliography

@inproceedings{ijcai2024p252,
  title     = {Multi-Modal Sarcasm Detection Based on Dual Generative Processes},
  author    = {Ma, Huiying and He, Dongxiao and Wang, Xiaobao and Jin, Di and Ge, Meng and Wang, Longbiao},
  booktitle = {Proceedings of the Thirty-Third International Joint Conference on
               Artificial Intelligence, {IJCAI-24}},
  publisher = {International Joint Conferences on Artificial Intelligence Organization},
  editor    = {Kate Larson},
  pages     = {2279--2287},
  year      = {2024},
  month     = {8},
  note      = {Main Track},
  doi       = {10.24963/ijcai.2024/252},
  url       = {https://doi.org/10.24963/ijcai.2024/252},
}

@article{zampieri2023offensive,
  title={Offensive language identification with multi-task learning},
  author={Zampieri, Marcos and Ranasinghe, Tharindu and Sarkar, Diptanu and Ororbia, Alex},
  journal={Journal of Intelligent Information Systems},
  volume={60},
  number={3},
  pages={613--630},
  year={2023},
  publisher={Springer}
}

@misc{lin2024vilapretrainingvisuallanguage,
      title={VILA: On Pre-training for Visual Language Models}, 
      author={Ji Lin and Hongxu Yin and Wei Ping and Yao Lu and Pavlo Molchanov and Andrew Tao and Huizi Mao and Jan Kautz and Mohammad Shoeybi and Song Han},
      year={2024},
      eprint={2312.07533},
      archivePrefix={arXiv},
      primaryClass={cs.CV},
      url={https://arxiv.org/abs/2312.07533}, 
}

@article{NEURIPS2022_960a172b,
  title={Flamingo: a visual language model for few-shot learning},
  author={Alayrac, Jean-Baptiste and Donahue, Jeff and Luc, Pauline and Miech, Antoine and Barr, Iain and Hasson, Yana and Lenc, Karel and Mensch, Arthur and Millican, Katherine and Reynolds, Malcolm and others},
  journal={Advances in neural information processing systems},
  volume={35},
  pages={23716--23736},
  year={2022}
}

@misc{yaghoobian2021sarcasmdetectioncomparativestudy,
      title={Sarcasm Detection: A Comparative Study}, 
      author={Hamed Yaghoobian and Hamid R. Arabnia and Khaled Rasheed},
      year={2021},
      eprint={2107.02276},
      archivePrefix={arXiv},
      primaryClass={cs.CL},
      url={https://arxiv.org/abs/2107.02276}, 
}

@misc{yang2023revilmretrievalaugmentedvisuallanguage,
      title={Re-ViLM: Retrieval-Augmented Visual Language Model for Zero and Few-Shot Image Captioning}, 
      author={Zhuolin Yang and Wei Ping and Zihan Liu and Vijay Korthikanti and Weili Nie and De-An Huang and Linxi Fan and Zhiding Yu and Shiyi Lan and Bo Li and Ming-Yu Liu and Yuke Zhu and Mohammad Shoeybi and Bryan Catanzaro and Chaowei Xiao and Anima Anandkumar},
      year={2023},
      eprint={2302.04858},
      archivePrefix={arXiv},
      primaryClass={cs.CV},
      url={https://arxiv.org/abs/2302.04858}, 
}

@article{helal2024contextual,
  author    = {Nivin A. Helal and Ahmed Hassan and Nagwa L. Badr and Yasmine M. Afify},
  title     = {A contextual-based approach for sarcasm detection},
  journal   = {Scientific Reports},
  volume    = {14},
  number    = {1},
  pages     = {15415},
  year      = {2024},
  doi       = {10.1038/s41598-024-65217-8},
  url       = {https://www.nature.com/articles/s41598-024-65217-8}
}

@Article{info13080399,
AUTHOR = {Băroiu, Alexandru-Costin and Trăușan-Matu, Ștefan},
TITLE = {Automatic Sarcasm Detection: Systematic Literature Review},
JOURNAL = {Information},
VOLUME = {13},
YEAR = {2022},
NUMBER = {8},
ARTICLE-NUMBER = {399},
URL = {https://www.mdpi.com/2078-2489/13/8/399},
ISSN = {2078-2489},
DOI = {10.3390/info13080399}
}

@misc{desai2021nice,
      title={Nice perfume. How long did you marinate in it? Multimodal Sarcasm Explanation}, 
      author={Poorav Desai and Tanmoy Chakraborty and Md Shad Akhtar},
      year={2021},
      eprint={2112.04873},
      archivePrefix={arXiv},
      primaryClass={cs.CL}
}

@inproceedings{qin-etal-2023-mmsd2,
    title = "{MMSD}2.0: Towards a Reliable Multi-modal Sarcasm Detection System",
    author = "Qin, Libo  and
      Huang, Shijue  and
      Chen, Qiguang  and
      Cai, Chenran  and
      Zhang, Yudi  and
      Liang, Bin  and
      Che, Wanxiang  and
      Xu, Ruifeng",
    booktitle = "Findings of the Association for Computational Linguistics: ACL 2023",
    month = jul,
    year = "2023",
    address = "Toronto, Canada",
    publisher = "Association for Computational Linguistics",
    url = "https://aclanthology.org/2023.findings-acl.689",
    pages = "10834--10845",
}

@inproceedings{yue-etal-2024-sarcnet,
    title = "{S}arc{N}et: A Multilingual Multimodal Sarcasm Detection Dataset",
    author = "Yue, Tan  and
      Shi, Xuzhao  and
      Mao, Rui  and
      Hu, Zonghai  and
      Cambria, Erik",
    editor = "Calzolari, Nicoletta  and
      Kan, Min-Yen  and
      Hoste, Veronique  and
      Lenci, Alessandro  and
      Sakti, Sakriani  and
      Xue, Nianwen",
    booktitle = "Proceedings of the 2024 Joint International Conference on Computational Linguistics, Language Resources and Evaluation (LREC-COLING 2024)",
    month = may,
    year = "2024",
    address = "Torino, Italia",
    publisher = "ELRA and ICCL",
    url = "https://aclanthology.org/2024.lrec-main.1248/",
    pages = "14325--14335"
}

@inproceedings{li2023blip2,
      title={{BLIP-2:} Bootstrapping Language-Image Pre-training with Frozen Image Encoders and Large Language Models}, 
      author={Junnan Li and Dongxu Li and Silvio Savarese and Steven Hoi},
      year={2023},
      booktitle={ICML},
}

@inproceedings{
dai2023instructblip,
title={Instruct{BLIP}: Towards General-purpose Vision-Language Models with Instruction Tuning},
author={Wenliang Dai and Junnan Li and Dongxu Li and Anthony Tiong and Junqi Zhao and Weisheng Wang and Boyang Li and Pascale Fung and Steven Hoi},
booktitle={Thirty-seventh Conference on Neural Information Processing Systems},
year={2023},
url={https://openreview.net/forum?id=vvoWPYqZJA}
}

@misc{awadalla2023openflamingoopensourceframeworktraining,
      title={OpenFlamingo: An Open-Source Framework for Training Large Autoregressive Vision-Language Models}, 
      author={Anas Awadalla and Irena Gao and Josh Gardner and Jack Hessel and Yusuf Hanafy and Wanrong Zhu and Kalyani Marathe and Yonatan Bitton and Samir Gadre and Shiori Sagawa and Jenia Jitsev and Simon Kornblith and Pang Wei Koh and Gabriel Ilharco and Mitchell Wortsman and Ludwig Schmidt},
      year={2023},
      eprint={2308.01390},
      archivePrefix={arXiv},
      primaryClass={cs.CV},
      url={https://arxiv.org/abs/2308.01390}, 
}

@inproceedings{liu2023llava,
    author      = {Liu, Haotian and Li, Chunyuan and Wu, Qingyang and Lee, Yong Jae},
    title       = {Visual Instruction Tuning},
    booktitle   = {NeurIPS},
    year        = {2023}
  }

@article{beyer2024paligemma,
    title={{PaliGemma: A versatile 3B VLM for transfer}},
    author={Lucas Beyer* and Andreas Steiner* and André Susano Pinto* and Alexander Kolesnikov* and Xiao Wang* and Daniel Salz and Maxim Neumann and Ibrahim Alabdulmohsin and Michael Tschannen and Emanuele Bugliarello and Thomas Unterthiner and Daniel Keysers and Skanda Koppula and Fangyu Liu and Adam Grycner and Alexey Gritsenko and Neil Houlsby and Manoj Kumar and Keran Rong and Julian Eisenschlos and Rishabh Kabra and Matthias Bauer and Matko Bošnjak and Xi Chen and Matthias Minderer and Paul Voigtlaender and Ioana Bica and Ivana Balazevic and Joan Puigcerver and Pinelopi Papalampidi and Olivier Henaff and Xi Xiong and Radu Soricut and Jeremiah Harmsen and Xiaohua Zhai*},
    year={2024},
    journal={arXiv preprint arXiv:2407.07726}
}

@article{Qwen-VL,
  title={Qwen-VL: A Frontier Large Vision-Language Model with Versatile Abilities},
  author={Bai, Jinze and Bai, Shuai and Yang, Shusheng and Wang, Shijie and Tan, Sinan and Wang, Peng and Lin, Junyang and Zhou, Chang and Zhou, Jingren},
  journal={arXiv preprint arXiv:2308.12966},
  year={2023}
}

@inproceedings{radford2021clip,
  title     = {Learning Transferable Visual Models From Natural Language Supervision},
  author    = {Radford, Alec and Kim, Jong Wook and Hallacy, Chris and Ramesh, Aditya and Goh, Gabriel and Agarwal, Sandhini and Sastry, Girish and Askell, Amanda and Mishkin, Pamila and Clark, Jack and Krueger, Gretchen and Sutskever, Ilya},
  booktitle = {Proceedings of the 38th International Conference on Machine Learning (ICML)},
  pages     = {8748--8763},
  year      = {2021},
  publisher = {PMLR}
}

@inproceedings{hessel2021clipscore,
  title     = {CLIPScore: A Reference-free Evaluation Metric for Image Captioning},
  author    = {Hessel, Jack and Holtzman, Ari and Forbes, Maxwell and Le Bras, Ronan and Choi, Yejin},
  booktitle = {Proceedings of the 2021 Conference on Empirical Methods in Natural Language Processing (EMNLP)},
  pages     = {7514--7528},
  year      = {2021},
  publisher = {Association for Computational Linguistics},
  doi       = {10.18653/v1/2021.emnlp-main.595}
}

@inproceedings{hertz2023delta,
  title     = {Delta Denoising Score},
  author    = {Hertz, Amir and Perel, Or and Giryes, Raja and Bermano, Amit H},
  booktitle = {Proceedings of the IEEE/CVF International Conference on Computer Vision (ICCV)},
  pages     = {17721--17730},
  year      = {2023},
  doi       = {10.1109/ICCV51070.2023.01623}
}

@article{gemma_2025,
    title={Gemma 3},
    url={https://goo.gle/Gemma3Report},
    publisher={Kaggle},
    author={Gemma Team},
    year={2025}
}

@inproceedings{chaudhari2017literature,
  title={Literature survey of sarcasm detection},
  author={Chaudhari, Pranali and Chandankhede, Chaitali},
  booktitle={2017 International conference on wireless communications, signal processing and networking (WiSPNET)},
  pages={2041--2046},
  year={2017},
  organization={IEEE}
}

@article{joshi2017automatic,
  title={Automatic sarcasm detection: A survey},
  author={Joshi, Aditya and Bhattacharyya, Pushpak and Carman, Mark J},
  journal={ACM Computing Surveys (CSUR)},
  volume={50},
  number={5},
  pages={1--22},
  year={2017},
  publisher={ACM New York, NY, USA}
}

@inproceedings{badlani2019ensemble,
  title={An ensemble of humour, sarcasm, and hate speechfor sentiment classification in online reviews},
  author={Badlani, Rohan and Asnani, Nishit and Rai, Manan},
  booktitle={Proceedings of the 5th Workshop on Noisy User-generated Text (W-NUT 2019)},
  pages={337--345},
  year={2019}
}

@inproceedings{maynard2014cares,
  title={Who cares about sarcastic tweets? investigating the impact of sarcasm on sentiment analysis},
  author={Maynard, Diana G and Greenwood, Mark A},
  booktitle={Lrec 2014 proceedings},
  year={2014},
  organization={ELRA}
}

@inproceedings{frenda2018role,
  title={The role of sarcasm in hate speech. A multilingual perspective},
  author={Frenda, Simona and others},
  booktitle={Proceedings of the doctoral symposium of the xxxivinternational conference of the spanish society for natural language processing (sepln 2018)},
  pages={13--17},
  year={2018},
  organization={Lloret, E.; Saquete, E.; Mart ́{\i}nez-Barco, P.; Moreno, I.}
}

@article{rothermich2021change,
  title={Change in humor and sarcasm use based on anxiety and depression symptom severity during the COVID-19 pandemic},
  author={Rothermich, Kathrin and Ogunlana, Ayotola and Jaworska, Natalia},
  journal={Journal of psychiatric research},
  volume={140},
  pages={95--100},
  year={2021},
  publisher={Elsevier}
}

@inproceedings{verma2021techniques,
  title={Techniques of sarcasm detection: A review},
  author={Verma, Palak and Shukla, Neha and Shukla, AP},
  booktitle={2021 International Conference on Advance Computing and Innovative Technologies in Engineering (ICACITE)},
  pages={968--972},
  year={2021},
  organization={IEEE}
}

@inproceedings{salini2023sarcasm,
  title={Sarcasm detection: A systematic review of methods and approaches},
  author={Salini, Yalamanchili and HariKiran, J},
  booktitle={2023 3rd International Conference on Smart Data Intelligence (ICSMDI)},
  pages={15--22},
  year={2023},
  organization={IEEE}
}

@inproceedings{cai2019multi,
  title={Multi-modal sarcasm detection in twitter with hierarchical fusion model},
  author={Cai, Yitao and Cai, Huiyu and Wan, Xiaojun},
  booktitle={Proceedings of the 57th annual meeting of the association for computational linguistics},
  pages={2506--2515},
  year={2019}
}

@inproceedings{mfarabi,
  title     = {A Survey of Multimodal Sarcasm Detection},
  author    = {Farabi, Shafkat and Ranasinghe, Tharindu and Kanojia, Diptesh and Kong, Yu and Zampieri, Marcos},
  booktitle = {Proceedings of the Thirty-Third International Joint Conference on
               Artificial Intelligence, {IJCAI-24}},
  publisher = {International Joint Conferences on Artificial Intelligence Organization},
  editor    = {Kate Larson},
  pages     = {8020--8028},
  year      = {2024},
  month     = {8},
  note      = {Survey Track},
  doi       = {10.24963/ijcai.2024/887},
  url       = {https://doi.org/10.24963/ijcai.2024/887},
}

@inproceedings{schifanella2016detecting,
  title={Detecting sarcasm in multimodal social platforms},
  author={Schifanella, Rossano and De Juan, Paloma and Tetreault, Joel and Cao, Liangliang},
  booktitle={Proceedings of the 24th ACM international conference on Multimedia},
  pages={1136--1145},
  year={2016}
}

@inproceedings{das2018sarcasmB,
  title={Sarcasm detection on flickr using a cnn},
  author={Das, Dipto and Clark, Anthony J},
  booktitle={Proceedings of the 2018 international conference on computing and big data},
  pages={56--61},
  year={2018}
}

@inproceedings{pan2020modeling,
  title={Modeling intra and inter-modality incongruity for multi-modal sarcasm detection},
  author={Pan, Hongliang and Lin, Zheng and Fu, Peng and Qi, Yatao and Wang, Weiping},
  booktitle={Findings of the Association for Computational Linguistics: EMNLP 2020},
  pages={1383--1392},
  year={2020}
}

@inproceedings{xu2020reasoning,
  title={Reasoning with multimodal sarcastic tweets via modeling cross-modality contrast and semantic association},
  author={Xu, Nan and Zeng, Zhixiong and Mao, Wenji},
  booktitle={Proceedings of the 58th annual meeting of the association for computational linguistics},
  pages={3777--3786},
  year={2020}
}

@inproceedings{liang2021multia,
  title={Multi-modal sarcasm detection with interactive in-modal and cross-modal graphs},
  author={Liang, Bin and Lou, Chenwei and Li, Xiang and Gui, Lin and Yang, Min and Xu, Ruifeng},
  booktitle={Proceedings of the 29th ACM international conference on multimedia},
  pages={4707--4715},
  year={2021}
}

@inproceedings{liang2022multib,
  title={Multi-modal sarcasm detection via cross-modal graph convolutional network},
  author={Liang, Bin and Lou, Chenwei and Li, Xiang and Yang, Min and Gui, Lin and He, Yulan and Pei, Wenjie and Xu, Ruifeng},
  booktitle={Proceedings of the 60th Annual Meeting of the Association for Computational Linguistics (Volume 1: Long Papers)},
  pages={1767--1777},
  year={2022}
}

@inproceedings{wang-etal-2020-building,
    title = "Building a Bridge: A Method for Image-Text Sarcasm Detection Without Pretraining on Image-Text Data",
    author = "Wang, Xinyu  and
      Sun, Xiaowen  and
      Yang, Tan  and
      Wang, Hongbo",
    editor = "Castellucci, Giuseppe  and
      Filice, Simone  and
      Poria, Soujanya  and
      Cambria, Erik  and
      Specia, Lucia",
    booktitle = "Proceedings of the First International Workshop on Natural Language Processing Beyond Text",
    month = nov,
    year = "2020",
    address = "Online",
    publisher = "Association for Computational Linguistics",
    url = "https://aclanthology.org/2020.nlpbt-1.3/",
    doi = "10.18653/v1/2020.nlpbt-1.3",
    pages = "19--29"
}

@article{lin2023goatbench,
  title={GOAT-Bench: Safety Insights to Large Multimodal Models through Meme-Based Social Abuse},
  author={Lin, Hongzhan and Luo, Ziyang and Wang, bo and Yang, Ruichao and Ma, Jing},
  journal={arXiv preprint arXiv:2401.01523},
  year={2024}
}

@article{yang2023mm,
  title={Mm-bigbench: Evaluating multimodal models on multimodal content comprehension tasks},
  author={Yang, Xiaocui and Wu, Wenfang and Feng, Shi and Wang, Ming and Wang, Daling and Li, Yang and Sun, Qi and Zhang, Yifei and Fu, Xiaoming and Poria, Soujanya},
  journal={arXiv preprint arXiv:2310.09036},
  year={2023}
}

@inproceedings{tang-etal-2024-leveraging,
    title = "Leveraging Generative Large Language Models with Visual Instruction and Demonstration Retrieval for Multimodal Sarcasm Detection",
    author = "Tang, Binghao  and
      Lin, Boda  and
      Yan, Haolong  and
      Li, Si",
    editor = "Duh, Kevin  and
      Gomez, Helena  and
      Bethard, Steven",
    booktitle = "Proceedings of the 2024 Conference of the North American Chapter of the Association for Computational Linguistics: Human Language Technologies (Volume 1: Long Papers)",
    month = jun,
    year = "2024",
    address = "Mexico City, Mexico",
    publisher = "Association for Computational Linguistics",
    url = "https://aclanthology.org/2024.naacl-long.97/",
    doi = "10.18653/v1/2024.naacl-long.97",
    pages = "1732--1742"
}

@inproceedings{ranasinghe2021mudes,
  title={MUDES: Multilingual Detection of Offensive Spans},
  author={Ranasinghe, Tharindu and Zampieri, Marcos},
  booktitle={Proceedings of NAACL},
  pagesNO={144--152},
  year={2021}
}

@article{srirag2024besstie,
  title={Besstie: A benchmark for sentiment and sarcasm classification for varieties of english},
  author={Srirag, Dipankar and Joshi, Aditya and Painter, Jordan and Kanojia, Diptesh},
  journal={arXiv preprint arXiv:2412.04726},
  year={2024}
}

@article{song2025large,
  title={Large Language Models for Subjective Language Understanding: A Survey},
  author={Song, Changhao and Zhang, Yazhou and Gao, Hui and Yao, Ben and Zhang, Peng},
  journal={arXiv preprint arXiv:2508.07959},
  year={2025}
}

@inproceedings{dmonte2024evaluation,
  title={An evaluation of large language models in financial sentiment analysis},
  author={Dmonte, Alphaeus and Ko, Eunmi and Zampieri, Marcos},
  booktitle={Proceedings of BigData},
  pagesNO={4869--4874},
  year={2024},
  organizationNO={IEEE}
}

@inproceedings{ranasinghe2020multilingual,
  title={Multilingual Offensive Language Identification with Cross-lingual Embeddings},
  author={Ranasinghe, Tharindu and Zampieri, Marcos},
  booktitle={Proceedings of EMNLP},
  pagesNO={5838--5844},
  year={2020}
}

\clearpage

\appendix
\section{Appendix}
\label{sec:appendix}

\subsection{Experimental Details}
\label{app:exp}
The models used in our experiments are open-source and sourced from the Hugging Face model repository.  All models are used \texttt{"out‑of‑the‑box”} (no gradient updates), loaded at their public checkpoints, and evaluated under identical hardware and preprocessing conditions. We utilized the transformers library to load both the models and their corresponding processors. Most models were loaded in 8-bit precision with FlashAttention enabled to optimize memory and computation efficiency. During inference, we used each model’s default tokenizer and processing pipeline for input encoding, text generation, and output parsing.

For the few-shot learning setting (1–3 examples), we randomly sampled positive examples from the MUSE dataset—since it contains explanations—and negative examples from the MMSD2.0 dataset. We then constructed a unified prompt for few-shot learning, as illustrated in Listing~\ref{lst:prompt_md_single}. The generated outputs were parsed using regular expressions, typically matching responses that begin with "Yes" or "No". Outputs that did not match the parsing pattern were logged but excluded from the reported scores in Table~\ref{tab:results_generated}. 

\textbf{Compute Resources}: 
All experiments were conducted on the high-performance computing cluster. We utilized a single compute node equipped with two NVIDIA A100 GPUs (40GB/80GB each), supported by 20GB and 40GB of system RAM. The approximate processing time for each model to handle a batch of 3000 examples (with a batch size of 1) ranged from 2 hours to 2 days, depending on the model and the task.

\textbf{Global Prompt}:
This can be found in in \texttt{params} file on the GitHub repository made available after acceptance of this paper.
\begin{lstlisting}[style=basiccode, caption={Unified Instruction for MSD}, label={lst:expl_unified_inst}]
"You are an image-caption sarcasm detector.\n"
"Sarcasm can arise when the text means something different than the image.\n"
"For each input, decide whether the caption is used sarcastically given the image.\n"
"- If sarcastic, output exactly:\n"
"1. Yes"
"\n2. On the next line, A single non-sarcastic rephrase of the caption (no extra text)\n"
"- If not sarcastic, output exactly:\n"
"1. No"
"\n\nIMPORTANT:\n"
"- Use capital "Y" in Yes and capital "N" in No and nothing else (no punctuation, no bullets).\n"
"DO NOT repeat or echo any part of this instruction or the caption.\n"
"Your entire response must be either:'No' or 'Yes' \n <rephrase>`\n"




\end{lstlisting}

These paradigms were adapted for each model's specific input requirements:
\begin{itemize}
    \item \textbf{LLaVA}: Employed a chat format \texttt{[INST] <..> [/INST]} (\texttt{USER}/\texttt{ASSISTANT}) with \texttt{<image>} placeholders; the processor handled multiple PIL images (examples + target) corresponding to tokens in a single prompt.
    
    \item \textbf{OpenFlamingo}: Required pre-processing images into tensors. A single text prompt interleaved \texttt{<image>} and text segments (separated by \texttt{<|endofchunk|>}), while the corresponding image tensors were passed separately as the \texttt{vision\_x} input.
    
    \item \textbf{InstructBlip \& Blip2}:  Processed a single target image per inference. Few-shot examples were provided textually within the prompt string as Q\&A pairs preceding the final query.

    \item \textbf{PaliGemma}: \texttt{<image>} placeholders within a single, interleaved prompt string; the processor handled multiple PIL images corresponding to tokens.
    \item \textbf{Qwen-VL} Embedded image file paths directly within the text prompt using \texttt{<img>path/to/image.jpg</img>} tags; the tokenizer handled image reference and loading internally. Few-shot examples also used this path embedding format within the prompt prefix.

    \item \textbf{Gemma3}: The inputs were structured in a chat-style format comprising an instruction, a user query, and an image represented as a PIL object. In the few-shot setting, demonstration examples were appended directly before the user's query to provide context for the model.
\end{itemize}

\textbf{Inference and Parsing}
\noindent
For each prompt, we generate a single output sequence with beam search (beam size=3 ) and a maximum of 100-256 generated tokens; for most models. We then parse the model’s reply via a lightweight regex to extract.

\smallskip
\texttt{Classification}: \texttt{Yes} $\rightarrow$ sarcastic  \texttt{No} $\rightarrow$ non-sarcastic\\
\texttt{Explanation}: A VLM generated explanation for sarcastic image-caption pairs.

Any unparsable or generic fallback (\textit{“I’m not trained…”}) is logged as an abstention. The unexpected results do not contribute to the final result.

\subsection{Hyperparameters}
\begin{table}[H]
\small
\centering
\caption{Inference-time hyperparameters}
\label{tab:hyperparams}
\begin{tabular}{@{}ll@{}}
\toprule
\textbf{Hyperparameter} & \textbf{Value} \\
\midrule
Batch size          & 1         \\
Example size        & 0-3      \\
Max tokens          & 100-256  \\
Num beams           & 3         \\
Quantization        & FP8, FP16 \\
Flash attention     & True      \\
Input padding       & True      \\
Input truncation    & True      \\
Temperature         & 0.7       \\
\bottomrule
\end{tabular}
\end{table}

The hyperparameters were selected to balance model compatibility and inference efficiency; for instance, Gemini, we limited our experiments to zero- and one-shot settings to avoid truncation of longer input sequences, thereby preserving the full contextual information necessary for reliable evaluation.

\end{document}